\documentclass[runningheads]{llncs}
\usepackage{eccv}
\usepackage{eccvabbrv}

\usepackage[utf8]{inputenc}
\usepackage[T1]{fontenc}
\usepackage{graphicx}
\usepackage{booktabs}
\usepackage[table,dvipsnames]{xcolor}
\usepackage{colortbl}
\usepackage{amsmath,amsfonts,bm}
\usepackage{nicefrac}
\usepackage{microtype}
\usepackage{multirow}
\usepackage{multicol}
\usepackage{makecell}
\usepackage{newfloat}
\usepackage{listings}
\usepackage{tcolorbox}
\usepackage{arydshln}
\usepackage{algorithm}
\usepackage{algpseudocode}
\usepackage{url}

\usepackage[square,numbers,sort&compress]{natbib}

\usepackage{caption}

\usepackage[pagebackref,breaklinks,colorlinks]{hyperref}
\hypersetup{
  colorlinks=true,
  citecolor=blue,
  linkcolor=brown,
  urlcolor=black,
  pdfborder={0 0 0}
}

\author{
Yun Wang\inst{1}$^{*}$,
Long Zhang\inst{2}$^{*}$,
Jingren Liu\inst{3},
Jiaqi Yan\inst{4},
Zhanjie Zhang\inst{5},\\
Jiahao Zheng\inst{1},
Ao Ma \inst{6},
Run Ling\inst{6},
Xun Yang\inst{2},
Dapeng Wu\inst{1},\\
Xiangyu Chen\inst{7}$^\ddagger$,
and Xuelong Li\inst{7}
}

\institute{
    \textsuperscript{\rm 1} City University of Hong Kong, 
    \textsuperscript{\rm 2} University of Science and Technology of China, 
    \textsuperscript{\rm 3} Tianjin University, 
    \textsuperscript{\rm 4} Nanjing University,
    \textsuperscript{\rm 5} Zhejiang University,
     \textsuperscript{\rm 6} JD.com,\\
    \textsuperscript{\rm 7} The Institute of Artificial Intelligence (TeleAI), China Telecom \\
    \{ywang3875-c, jhzheng4-c\}@my.cityu.edu.hk,
    dragonzhang@mail.ustc.edu.cn,
    jiaqi\_yan@smail.nju.edu.cn,
    cszzj@zju.edu.cn, 
    xyang21@ustc.edu.cn,
    dapengwu@cityu.edu.hk,
    \{chengxuyuangg, chxy95\}@gmail.com, 
    xuelong\_li@ieee.org \\
}

\definecolor{MyForestGreen}{HTML}{228B22}

\begin{document}
\title{Video-EM: Event-Centric Episodic Memory for Long-Form Video Understanding}
\titlerunning{Abbreviated paper title}
\maketitle
\begingroup
\renewcommand\thefootnote{*}
\footnotetext{Equal contribution.}

\renewcommand\thefootnote{$\ddagger$}
\footnotetext{Corresponding author.}
\endgroup

\begin{abstract}
Video Large Language Models (Video-LLMs) have shown strong video understanding, yet their application to long-form videos remains constrained by limited context windows. 
A common workaround is to compress long videos into a handful of representative frames via retrieval or summarization. However, most existing pipelines score frames in isolation, implicitly assuming that frame-level saliency is sufficient for downstream reasoning. This often yields redundant selections, fragmented temporal evidence, and weakened narrative grounding for long-form video question answering.
We present \textbf{Video-EM}, a training-free, event-centric episodic memory framework that reframes long-form VideoQA as \emph{episodic event construction} followed by \emph{memory refinement}. 
Instead of treating retrieved keyframes as independent visuals, Video-EM employs an LLM as an active memory agent to orchestrate off-the-shelf tools: it first localizes query-relevant moments via multi-grained semantic matching, then groups and segments them into temporally coherent events, and finally encodes each event as a grounded episodic memory with explicit temporal indices and spatio-temporal cues (capturing \emph{when}, \emph{where}, \emph{what}, and involved entities). 
To further suppress verbosity and noise from imperfect upstream signals, Video-EM integrates a reasoning-driven self-reflection loop that iteratively verifies evidence sufficiency and cross-event consistency, removes redundancy, and adaptively adjusts event granularity. The outcome is a compact yet reliable \emph{event timeline}---a minimal but sufficient episodic memory set that can be directly consumed by existing Video-LLMs without additional training or architectural changes.
Extensive experiments on long video understanding benchmarks demonstrate that Video-EM achieves highly competitive accuracy with substantially fewer frames than strong retrieval-based keyframe baselines.
\end{abstract}

\section{Introduction}
The rapid advancement of Video Large Language Models (Video-LLMs) has achieved remarkable progress in video understanding~\citep{2021vlm,zhang2025video}, particularly in video question answering~\cite{zhang2023video}, demonstrating strong potential for modeling real-world scenarios~\cite{song2025towards,zhang2025enhancing}. However, as video content moves from minutes to hours-long sequences, the limited context window of Video-LLMs becomes a bottleneck for long-form video understanding, making it difficult to preserve long-horizon evidence and coherent narratives.

\begin{figure*}[t]
\centering
\includegraphics[width=1\linewidth]{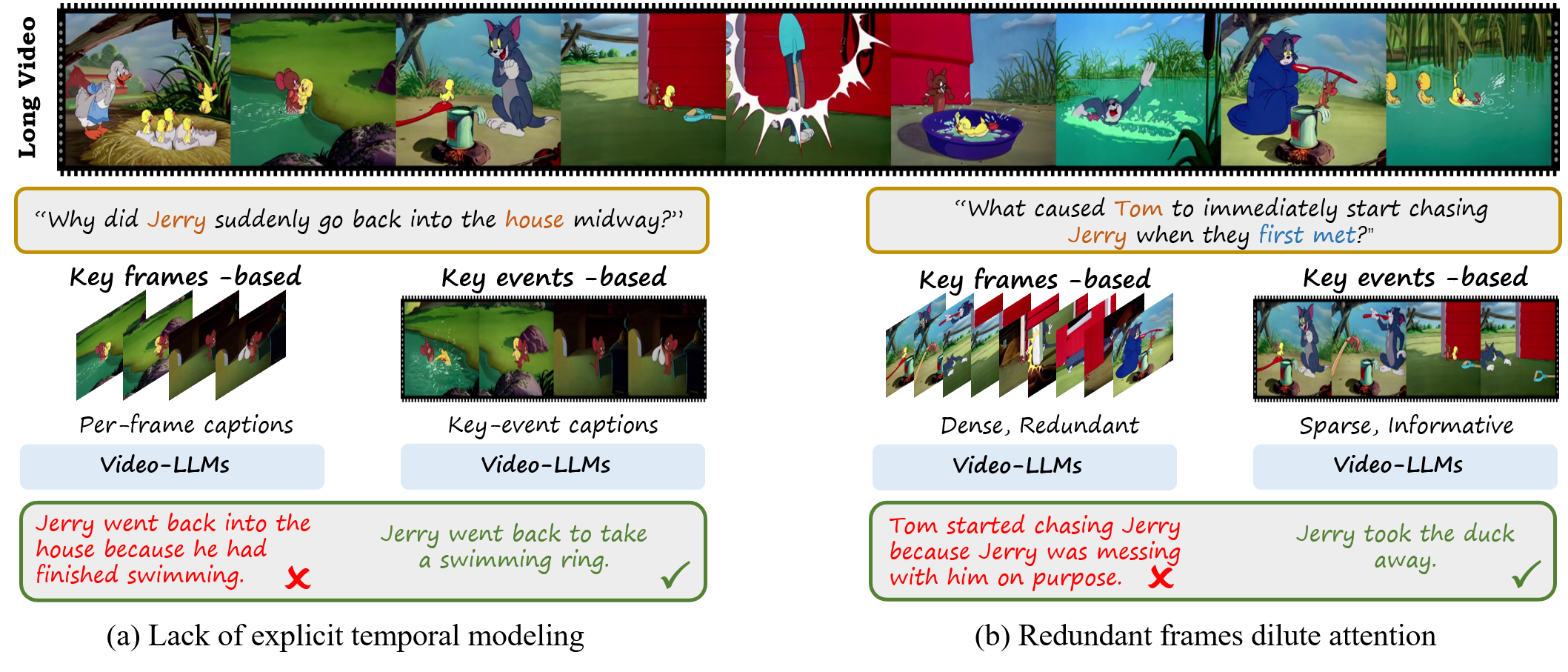}
    \vspace{-0.6cm}
    \caption{Illustration of the limitation of existing training‑free keyframe sampling methods. (a) Isolated frames break temporal continuity and weaken event narratives. (b) Redundant frames waste context and dilute key cues, hurting performance.}

    \label{fig:teaser}
    \vspace{-0.4cm}
\end{figure*}

To bridge this gap, a growing line of work studies training-free frame sampling and retrieval strategies for long-form video understanding~\cite{Liu_2025_CVPR,yu2024frame,tang2025adaptive,wang2024videotree,zhu2025focus}. These methods compress a long video into a small set of query-relevant keyframes, reducing the problem to static text-image matching. While effective in practice, this frame-centric formulation introduces two critical limitations for narrative reasoning.
First, selecting and captioning frames independently~\cite{wang2024videotree,liu2025bolt,rajan2025key} underexploits temporal coherence, obscuring scene transitions and weakening contextual continuity, both of which are essential for complex, multi-step questions over long-form videos.
Second, query-driven sampling frequently returns redundant frames due to repeated scenes or similar viewpoints in long videos. Such redundancy not only wastes the scarce context budget, but can also dilute salient cues and degrade downstream reasoning~\cite{fang2025threading,mdp3}, as illustrated in Figure~\ref{fig:teaser}.

In this work, we argue that long-form VideoQA~\cite{liu2025towards,wu2021towards} should be framed as \emph{event-centric} episodic memory construction rather than \emph{frame-centric} snapshot retrieval.
Humans achieve lifelong reasoning not by storing isolated snapshots, but by consolidating salient experiences into episodic memories—discrete events embedded in rich spatio-temporal contexts that support ``mental replay''~\footnote{\noindent In humans, episodic memory refers to the capacity to form, store, and consciously retrieve past events within their unique spatio-temporal contexts, effectively enabling ``mentally time travel'' to reexperience those moments.}~\cite{8421022,hampton2004episodic}. 
Inspired by this, we introduce Video-EM, a training-free agentic framework that reformulates long-form video understanding as a dynamic process of memory construction and refinement. 
Unlike static pipelines that treat frames as fixed and independent inputs, Video-EM employs the LLM as an active memory agent that orchestrates off-the-shelf tools to iteratively plan, group, and ground visual evidence.
In this way, Video-EM organizes them into temporally ordered events and encodes each event as an episodic memory with explicit temporal indices and grounding cues, preserving intra-event spatial grounding and inter-event temporal structure. This representation provides narrative grounding by capturing \emph{when} and \emph{where} events occur, \emph{what} happens, and \emph{which} objects are involved, enabling Video-LLMs to reason over long videos with compact evidence. However, such event memories may still be verbose,  Video-EM therefore incorporates a reasoning-driven reflection loop that adaptively prunes redundancy to target a {minimal yet sufficient} evidence set for each query.

Concretely, Video-EM operates in three stages: (i) it identifies relevant moments via multi-grained semantic matching; (ii) it proposes coherent events by grouping and segmenting around these moments to preserve context; and (iii) it refines these episodic memories via iterative tool orchestration. Through a self-reflection loop, the agent identifies and mitigates potential hallucinations or tool-level inconsistencies, ensuring that the resulting memories are not only grounded but also contextually reliable.
The outcome is a compact ``event timeline'', a minimal yet informative set of episodic memories that can be fed into existing Video-LLMs for accurate and efficient answering, without additional training or architectural modifications.
Overall, our contributions are threefold:

\begin{itemize}
    \item  We propose an \emph{event-centric} paradigm for long-form video understanding that leverages {episodic memory} as a structured, narrative-grounded representation, moving beyond conventional \emph{frame-centric} sampling.

    \item We introduce Video-EM, a training-free agentic framework where an LLM-based memory agent uses off-the-shelf tools to localize relevant moments, structure them into events, and refine grounded episodic memories into a minimal yet sufficient event timeline.

    \item  Extensive experiments on long video understanding benchmarks demonstrate that Video-EM consistently improves VideoQA performance with fewer frames, and is broadly compatible with mainstream Video-LLMs.
\end{itemize}

\section{Related Work}
\vspace{-0.1cm}
\subsection{Long-form Video Understanding}
Long-video understanding is challenging due to complex temporal dynamics and severe visual redundancy~\cite{qian2024streaming,zeng2024timesuite}. Prior efforts extend the effective context via long-context modeling or token compression~\cite{chen2024longvila,zhang2024long,team2023gemini,li2024llama,weng2024longvlm,MovieChat}, and employ query-aware keyframe sampling to reduce redundancy~\cite{tang2025adaptive,yu2024frame,Liu_2025_CVPR}. Meanwhile, CoT-based Video-LLMs enhance multi-step reasoning over visual inputs~\cite{vot,zhang2023simple,VideoAgent,13,wang2024videotree}. Despite progress, many pipelines still treat selected keyframes as independent evidence, underexploiting temporal dependencies and weakening narrative grounding.
Tool-augmented and agentic approaches further improve long-video reasoning by iteratively gathering and organizing auxiliary cues, e.g., VideoAgent~\cite{VideoAgent,13}, Video-RAG~\cite{luo2024video}, and WorldMM~\cite{yeo2025worldmm}. In contrast, Video-EM adopts an \emph{event-centric} formulation: it consolidates sparse observations into grounded episodic events and refines them into a minimal yet sufficient \emph{event timeline} that fits strict context budgets. We argue that effective long-video understanding requires temporally grounded reasoning that captures both fine-grained semantics and high-level narrative progression.

\vspace{-0.2cm}

\begin{figure*}[t]
    \centering
\includegraphics[width=1\linewidth]{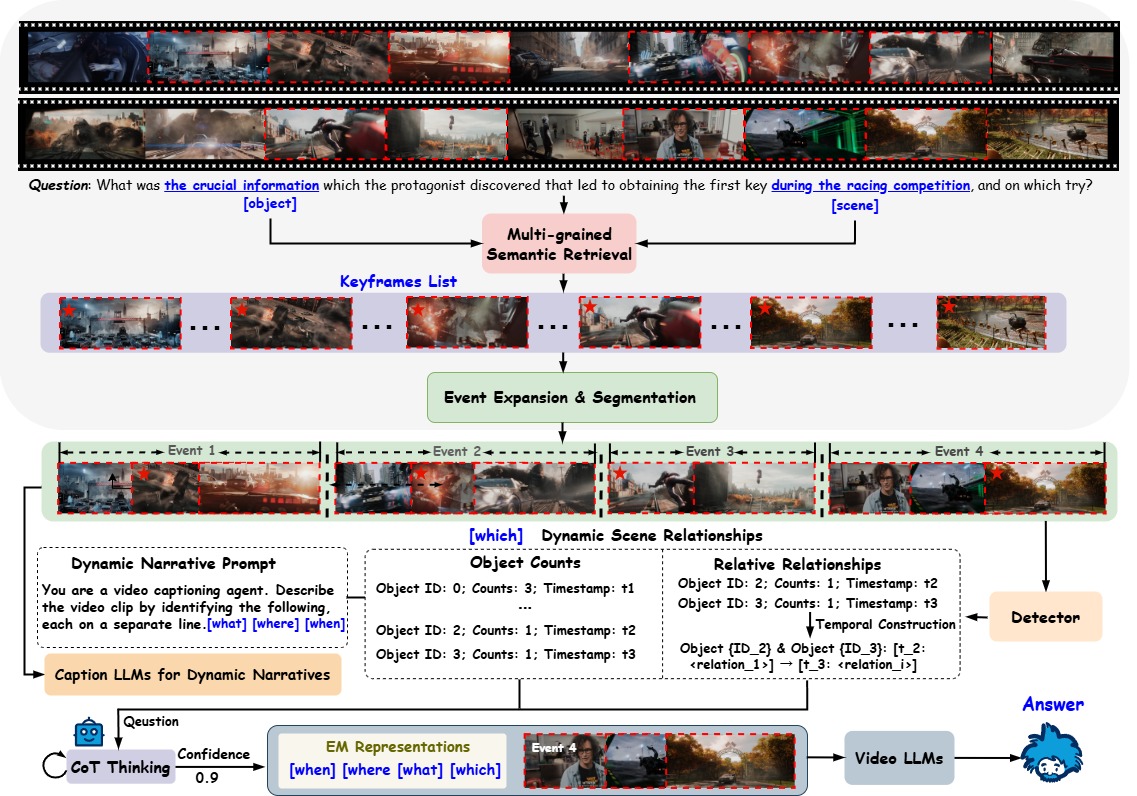}
\vspace{-0.5cm}
    \caption{The pipeline of the Video-EM framework consists of three steps: Key Event Selection, Episodic Memory Construction, and CoT-based Video Reasoning.}
    \label{fig:overview}
    \vspace{-0.4cm}
\end{figure*}

\subsection{Episodic Memory for VLM}
\vspace{-0.1cm}
Humans perform lifelong reasoning by organizing salient experiences into episodic memories grounded in spatio-temporal contexts~\cite{lee2007providing,moscovitch2016episodic,fountas2025humaninspired,ramakrishnan2023spotem}. Inspired by this, episodic memory has been adopted in embodied agents as hierarchical textual representations for long-horizon decision-making~\cite{8421022,9444649}. LifelongMemory~\cite{wang2024lifelongmemory} extends this idea to video understanding, yet its activity-centric memories emphasize \textit{what} while underrepresenting \textit{when/where} and involved objects, a gap also reflected in V-STaR~\cite{cheng2025vstar} where models struggle with precise spatio-temporal localization. Complementarily, GCAgent~\cite{yeo2025gcagent} introduces schematic and narrative episodic structures to better capture global context in long videos. In contrast, Video-EM prioritizes \emph{grounded} event memories that explicitly encode \emph{when}, \emph{where}, \emph{what}, and participating entities, yielding directly verifiable evidence under strict context budgets for long-form VideoQA.


\vspace{-0.2cm}
\section{Methodology}
\vspace{-0.2cm}

\textcolor{black}{We introduce Video-EM, a training-free agentic framework in which an LLM-based memory agent leverages off-the-shelf tools to (i) localize query-relevant moments via key event selection, (ii) construct temporally coherent events by organizing these moments into episodic units, and (iii) optimize grounded episodic memories through CoT-based iterative refinement, producing a minimal yet sufficient “event timeline” for downstream Video-LLMs (Figure~\ref{fig:overview}).}

\vspace{-0.2cm}
\subsection{Key  Event Selection}
Naive query-relevant moment localization often relies on coarse semantic matching, missing critical context due to limited semantic coverage, especially in long-video understanding~\cite{ma2022x}. To address this, our Key Event Selection module first performs multi-granular semantic retrieval to identify sparse yet informative candidate moments, and then applies event expansion and segmentation to obtain temporally contiguous moment segments around retrieved anchors. These segments serve as query-relevant evidence for subsequent episodic event construction.


\begin{table*}[t]
\begin{tcolorbox}[
        standard jigsaw,
        title=The workflow of Query Decomposition,
        opacityback=0,
        label=prompt_query_grounding,
        width=\textwidth,
    ]
\vspace{-0.1cm}
Analyze the following input question:

Question: \texttt{<Question>} , Options: \texttt{<Options>} 

\vspace{0.1cm}
\textbf{Process}: Key Object/Scene Identification
\vspace{0.1cm}

\quad\textbullet{} Extract key objects/scenes from the input query

\quad\textbullet{} Format: Key Objects: \texttt{obj1}, \texttt{obj2}, \texttt{obj3}

\quad\textbullet{} Format: Key Scenes: \texttt{Scene1}, \texttt{Scene2}, \texttt{Scene3}

\vspace{0.1cm}
\textbf{Output Rules}
\vspace{0.1cm}

\quad1. One line each for Key Objects and Scenes, starting with exact prefixes

\quad2. Separate items with a comma
    
\quad3. Never use markdown or natural language explanations
    
\quad4.  If you cannot identify any key objects or cue objects from the query provided, please output null

\vspace{0.1cm}
\textbf{Below is an example of the procedure:}
\vspace{0.1cm}

\quad Question: For “When does the person in red clothes appear with the dog at the fence?”

\quad Response:
    
\quad\quad Key Objects: \texttt{person, dog, red clothes}
        
\quad\quad Cue Scenes: \texttt{grassy\_area, fence}
        
\vspace{0.1cm}
\textbf{Format your response EXACTLY like this in two lines:}
\vspace{0.1cm}

\quad\quad Key Objects: \texttt{[object1, object2, object3]}
        
\quad\quad Key Scenes: \texttt{[Scene1, Scene2, Scene3]}
\vspace{-0.1cm}        
\end{tcolorbox}
\vspace{-0.6cm}  
\end{table*}

\noindent \textbf{Multi-grained Semantic Retrieval.}
Given a video $V=\left\{f_{i}\right\}^{N}_{i=1}$ consisting of $N$
frames, where $f_{i}$ is the $i$-th frame, our objective is to extract a set of representative keyframes that are most relevant to a given natural language query $q$.  
To achieve this, we first decompose $q$ into a multi-granular query set $Q = \left\{q_o, q_s, q_c\right\}$, where $q_o$ is the original query, $q_{s}$ captures object-level semantics (e.g., “apple”), and  $q_{c}$ captures scene-level context (e.g., “kitchen”), enabling robust and semantically aligned retrieval.
For each candidate frame $f_{i}$, we compute its similarity score as follows:
\begin{equation}
\vspace{-0.2cm}
Sim(f_{i}) = \sum\limits_{q_{j}\in Q} \omega_{q_{j}} \big( \tilde{\phi}_{I}(f_{i}) \cdot \tilde{\phi}_{T}(q_{j}) \big), 
\vspace{-0.1cm}
\end{equation}
where $\tilde{\phi}_{I}$ and $\tilde{\phi}_{T}$ denote the L2-normalized features extracted by the CLIP image and text encoders, respectively, such that their dot product directly computes the cosine similarity. The term $\omega_{q_{j}}$ represents the weight for each query component. To ensure scale invariance, the weights are normalized to sum to one (i.e., $\sum_{q_{j} \in Q} \omega_{q_{j}} = 1$), with $\omega_{q_o}$ assigned to the original query and the remaining weight equally distributed between the structural components ($\omega_{q_s}=\omega_{q_c}$).


\noindent \textbf{Event Expansion and Segmentation.}
\textcolor{black}{To obtain temporally and semantically coherent event representations, we first perform event expansion around each keyframe, followed by temporal event segmentation. Given a set of keyframes 
$V^{*}$, each keyframe is treated as a temporal anchor and is expanded along the video timeline to incorporate its surrounding context. Specifically, we extend the anchor keyframe bidirectionally and employ TransNetV2~\cite{soucek2024transnet} as a boundary-aware temporal model to infer potential event boundaries between adjacent frames. The expansion process continues as long as no boundary is detected; once TransNetV2 indicates the temporal discontinuity, the expansion in the corresponding direction is terminated. Through this procedure, each keyframe is augmented with contextually relevant frames, resulting in candidate event segments with enriched temporal continuity.}

\textcolor{black}{Based on the expanded keyframes, we further perform event segmentation by grouping temporally adjacent frames according to their timestamps. Event boundaries are determined by enforcing a minimum temporal gap  $\Delta t$ between consecutive events. Keyframes whose timestamps differ by less than $\Delta t$ are assigned to the same event $E_i$.
Formally, each event $E_i$ consists of a sequence of frames  $\left\{f^{E_{i}}_{j}\right\}^{n_{i}}_{j=1}$ where $n_i \leq N$ is the number of frames in $E_i$. 
The temporal distance between the last frame of $E_i$ and the first frame of $E_{i+1}$ must satisfy $|t_1^{(E_{i+1})} - t_{n_i}^{(E_{i})}| > \Delta t$. In our implementation, we empirically set the temporal gap $\Delta t$ to 3 frames. By iteratively applying this procedure to all keyframes, we obtain a set of temporally coherent events $\bm{\mathrm{E}}=\left\{E_{1}, E_{2}, \ldots, E_{M}\right\}$, where $M \leq K$.}

\vspace{-0.3cm}
\subsection{Grounded Episodic Memory Construction}
\vspace{-0.1cm}
After obtaining the key event set ${E}$, we promote each localized moment to an episodic unit by adaptively expanding its temporal window to recover query-relevant context that similarity-based retrieval may omit. 
This expansion helps preserve subtle transitions and causal cues, yielding temporally coherent clips. We then encode each expanded event as a grounded episodic memory, inspired by cognitive science~\cite{holland2011humans,hampton2004episodic}, explicitly capturing \emph{when}, \emph{where}, \emph{what}, and the \emph{involved entities}, thereby supporting modeling of dynamic spatio-temporal context.

\noindent \textbf{Dynamic Scene Narratives.} 
To comprehensively encode the \textit{when}, \textit{where}, and \textit{what} of each event $E_{i}$,we leverage the publicly available MLLM Qwen2.5-VL-7B~\cite{bai2025qwen2}, augmented with a custom reasoning prompt (Appendix \textcolor{blue}{Prompt C}). This model generates hierarchical, semantically rich summaries tightly anchored to temporal structure. In contrast to prior per-frame captioning methods, which often produce redundant or fragmented descriptions~\cite{wang2024videotree}, our approach delivers coherent, clip-level narratives that capture temporal evolution. Each summary $\bm{\mathrm{N}}^{scene}$ explicitly encodes the temporal position of the event (\textit{when}), its spatial context (\textit{where}), and the core actions and entities involved (\textit{what}). Beyond these scene-level summaries, complex video understanding further demands modeling dynamic interactions and evolving relationships among objects, which are critical for deeper situational comprehension and causal reasoning.

\noindent \textbf{Dynamic Scene Relationships.}
To address this, we construct event-centric scene representations that go beyond isolated object descriptions. Specifically, to explicitly capture the dynamic relationships of ``\textit{which objects}‘’ evolve and interact throughout each event, we encode fine-grained spatial and relational structures, denoted as $\bm{\mathrm{G}}^{scene}$.
\textcolor{black}{Leveraging a robust object detection framework~\cite{liu2024grounding}, we identify salient objects with high confidence (empirically set to 0.6) to model their dynamic spatial properties through two core components:}
\textit{(i)  Evolution of Object Counts $\bm{\mathrm{A}}_{cnt}$}:
This component characterizes the temporal evolution of object quantities by detecting their appearance and disappearance over time. It represents these dynamics in the following format:
``The count of {Object \{ID$_{1}$\} evolves as follows: $\{num_1\}$ at $t_1$, $\{num_2\}$ at $t_2$, ..., $\{num_n\}$ at $t_n$.”
{\textit{(ii) Evolution of Location Relationships} $\bm{\mathrm{A}}_{loc}$:
Captures temporal changes in pairwise spatial locations between objects, formatted as: `` Location evolution for Object \{ID$_{1}$\} \& Object  \{ID$_{2}$\}: [$t_{1}$: <relation$_{1}$>] →  [$t_{i}$: <relation$_{i}$>] → $\cdots$ →  [$t_{n}$: <relation$_{n}$>].''
These components combine into a structured scene relatiqonship representation: $\bm{\mathrm{G}}^{scene} =  \{\bm{\mathrm{A}}_{cnt}, \bm{\mathrm{A}}_{loc}\}$,
which provides an interpretable and structured representation of evolving scenes for video reasoning. 
In this way, the fine-grained spatio-temporal relationships captured by $\bm{\mathrm{G}}^{scene}$ and the high-level narratives in $\bm{\mathrm{N}}^{scene}$ complement each other, together forming a rich spatio-temporal episodic memory that supports effective, context-aware video reasoning.

\begin{algorithm}[t]
\small 
\vspace{0.1cm}
\caption{\small{The pseudo code of CoT with Evidence-Consistency}}
\label{alg:cot_ec}


\begin{tabular}[width=0.9\textwidth]{@{}ll@{}} 
    \multicolumn{2}{@{}l}{\textbf{Input:} Memory $\mathcal{M}$, Events $E$, Query $q$, Depth $d$} \\
    \multicolumn{2}{@{}l}{\textbf{Output:} Evidence, Ans \quad \textbf{Param:} $\tau, D_{max}$} \\
    \\[-0.25cm] 
    
    \textbf{1:} & \textbf{Function} \textsc{CoT\_Recurse}($\mathcal{M}, E, q, d$) \\
    
    \textbf{2:} & \quad $(r, conf, stat, \_, g) \gets \textsc{Predict}(\mathcal{M}, E, q)$ \\
    
    \textbf{3:} & \quad \textbf{if} $conf \ge \tau \land stat = \texttt{pass}$ \textbf{then} \\
    \textbf{4:} & \quad \quad \textbf{return} $\mathcal{M}, E$ \\
    \textbf{5:} & \quad \textbf{end if} \\
    
    \textbf{6:} & \quad \textbf{if} $d \ge D_{max}$ \textbf{or} $g=\texttt{coarsen}$ \textbf{then} \\
    \textbf{7:} & \quad \quad \textbf{return} \textsc{Construct}(\textsc{Merge}(E), \text{prompt}) \\
    \textbf{8:} & \quad \textbf{end if} \\
    
    \textbf{9:} & \quad $(stat, issue) \gets \textsc{Check}(\mathcal{M}, E, q)$ \\
    
    \textbf{10:} & \quad \textbf{if} $g = \texttt{refine}$ \textbf{then} \\
    \textbf{11:} & \quad \quad $[E^L, E^R] \gets \textsc{Split}(E, issue, q)$ \\
    \textbf{12:} & \quad \textbf{else} \quad $[E^L, E^R] \gets \textsc{Cluster}(E)$ \quad \textit{// Default Strategy} \\
    \textbf{14:} & \quad \textbf{end if} \\
    
    \textbf{15:} & \quad $\mathcal{M}^{L}, \mathcal{M}^{R} \gets \textsc{Construct}(E^{L/R}, \text{prompt})$ \\
    
    \textbf{16:} & \quad \textbf{if} $\textsc{Recurse}(\mathcal{M}^L, E^L, q, d{+}1)$ is valid \textbf{then} \\
    \textbf{17:} & \quad \quad \textbf{return} $\mathcal{M}^L, E^L$ \\
    \textbf{18:} & \quad \textbf{else} \\
    \textbf{19:} & \quad \quad \textbf{return} $\textsc{Recurse}(\mathcal{M}^R, E^R, q, d{+}1)$ \\
    \textbf{20:} & \quad \textbf{end if} \\
    
    \textbf{21:} & \textbf{End Function} \\
\end{tabular}
\end{algorithm}

\vspace{-0.4cm}
\subsection{Self-reflective Memory Refinement}
\vspace{-0.2cm}

\textcolor{black}{
Instead of feeding all retrieved keyframes/memories to downstream Video-LLMs, often introducing redundancy and diluting salient cues~\cite{Liu_2025_CVPR}, we adopt an \emph{event-centric episodic memory} representation and a reasoning-driven CoT refinement strategy to construct a \emph{minimal yet sufficient} evidence set per query (Alg.~\ref{alg:cot_ec}). 
Starting from event-level episodic memories that encode (i) intra-event spatial grounding (entities with attributes/relations) and (ii) inter-event temporal dynamics (ordering and transitions), the CoT agent iteratively \emph{selects}, \emph{filters}, and \emph{composes} them into a compact \emph{event timeline}. 
Each step is constrained to cite concrete evidence, specific events (with time spans), and referenced entities, thus preserving narrative continuity while avoiding frame-level redundancy.
}
\textcolor{black}{
Even with \emph{event-centric} organization, episodic memories may remain verbose or noisy due to imperfect upstream signals, and the evidence required by a query can vary in granularity. 
We therefore add an \emph{evidence sufficiency and consistency} verifier~\cite{wang2022self} to the loop: at each step, the agent checks whether the current timeline is sufficient to answer the query and whether evidence across events is self-consistent (e.g., attribute or temporal conflicts). 
If the check fails, the agent performs \emph{refine-or-fallback} within the same hierarchy—splitting a coarse event into finer sub-events for specific cues, or reverting to a higher-level summary for stable context—enabling backtracking and denoising. 
Details are provided in the Appendix.
}


\vspace{-0.5cm}
\section{Experiments}
\vspace{-0.1cm}
\subsection{Experimental Setup}

\noindent \textbf{Benchmarks.} 
We evaluate the performance of Video-EM in four popular benchmarks:
1) \textbf{Video-MME}~\cite{fu2025video}, comprising 2700 question-answer pairs, with an average video duration of 17 minutes.
2) \textbf{LVBench}~\cite{wang2024lvbench} is an hours-long benchmark with an average length of 4101 seconds  (68 minutes), 1549 question-answer pairs, and four multiple-choice options.
3) \textbf{HourVideo}~\cite{chandrasegaran2024hourvideo}, we use its dev set, including 50 videos with an average duration of 47.2 minutes, comprising 1182 high-quality, five-way multiple-choice questions.
4) \textbf{Egoschema}~\cite{mangalam2023egoschema} is a popular benchmark derived from Ego4D~\cite{Grauman_2022_CVPR}. It consists of 5-way multiple-choice questions based on videos, which are 180 seconds in length. We run ablations on its subset.

\noindent \textbf{Implementation Details.}
For benchmark evaluation, we sample frames at 1 fps (up to 1,024 per video) and use CLIP with a ViT-G backbone~\cite{radford2021learning} for multi-granular semantic retrieval. Our pipeline integrates four foundation models: TransNet v2~\cite{soucek2024transnet} for shot boundary detection, Grounding-DINO~\cite{liu2023grounding} for object detection, Qwen2.5-VL-7B~\cite{bai2025qwen2} for clip-level narrative modeling, and Qwen3-8B~\cite{qwen3} as the CoT reasoning agent. All experiments are run on an NVIDIA A100 GPU.

\begin{table*}[t]
    \centering
    \scriptsize
    \renewcommand{\arraystretch}{1.1} 
    \setlength{\tabcolsep}{3pt} 
    \resizebox{0.95\linewidth}{!}{ 
        \begin{tabular}{l c c c c c c c}
        \toprule 
        \multirow{2}{*}{Model} & \multirow{2}{*}{Venue} & \multirow{2}{*}{LLM Size} & \multirow{2}{*}{Frames} & Overall & Short & Medium & Long \\
        \cmidrule(lr){5-8} 
         & & & & 17 min & 1.3 min & 9 min & 41 min \\
        \midrule
        
        \rowcolor{gray!20} \multicolumn{8}{c}{\textit{Proprietary Models}} \\
        GPT-4o & - & - & 32 & 62.5 & 71.4 & 61.0 & 55.2 \\
        
        \midrule
        \rowcolor{gray!10} \multicolumn{8}{c}{\textit{Open-source Models}} \\
        Video-LLaVA & ArXiv'23 & 7B & 8 & 41.6 & 46.1 & 40.7 & 38.1 \\
        VideoLLaMA2 & ArXiv'24 & 7B & 8 & 47.9 & 56.0 & 45.4 & 42.1 \\
        LongVA & ArXiv'24 & 7B & 128 & 52.6 & 61.1 & 50.4 & 46.2 \\
        ShareGPT4Video & NIPS'24 & 8B & 16 & 43.6 & 53.6 & 39.3 & 37.9 \\
        LongVU & ICML'25 & 7B & 1fps & 60.6 & 64.7 & 58.2 & \textbf{59.5} \\
        Video-XL & CVPR'25 & 7B & 128 & 55.5 & 64.0 & 53.2 & 49.2 \\
        Video-XL-2 & Arxiv'25 & 8B & 1fps & 66.6 & - & - & - \\
        VideoChat2 & CVPR'24 & 7B & 16 & 43.8 & 52.8 & 39.4 & 39.2 \\
        Frame-Voyager & ICLR'25 & 8B & - & 57.5 & 67.3 & 56.3 & 48.9 \\ 
        
        \midrule
        \rowcolor{gray!5} \multicolumn{8}{c}{\textit{Training-free KeyFrame Selection Models}} \\
        VideoTree & CVPR'25 & - & 128 & 54.2 & - & - & - \\
        Qwen2-VL w/AKS & CVPR'25 & 7B & 32 & 59.9 & - & - & - \\
        Qwen2-VL w/BOLT$^\ddagger$ & CVPR'25 & 7B & 32 & 59.5 & 69.4 & 57.5 & 51.5 \\
        Qwen2-VL w/Q-Frame & ICCV'25 & 7B & 44 & 58.3 & 69.4 & 57.1 & 48.3 \\
        Qwen2-VL w/FOCUS & ICLR'26 & 7B & 32 & 59.7 & - & - & - \\
        \midrule
        Qwen2-VL$^\ddagger$ & ArXiv'24 & 7B & 32 & 56.9 & 68.7 & 53.3 & 48.8 \\
        \rowcolor{cyan!15} \textbf{Qwen2-VL w/Ours} & - & 7B & Avg 28 & \underline{61.2} & \underline{70.7} & \underline{58.8} & 52.3 \\
        \midrule
        Qwen2.5-VL$^\ddagger$ & ArXiv'25 & 7B & 32 & 59.2 & 70.6 & 56.7 & 50.6 \\
        \rowcolor{cyan!15} \textbf{Qwen2.5-VL w/Ours} & - & 7B & Avg 28 & \textbf{62.0} & \textbf{72.4} & \textbf{60.3} & \underline{53.4} \\
        \midrule
        LLaVA-OV$^\ddagger$ & ArXiv'24 & 7B & 32 & 56.5 & 67.8 & 54.5 & 47.2 \\
        \rowcolor{cyan!15} \textbf{LLaVA-OV w/Ours} & \textbf{-} & \textbf{7B} & \textbf{Avg 28} & \textbf{59.6} & \textbf{70.1} & \textbf{57.2} & \textbf{51.5} \\
        \midrule
        LLaVA-Video$^\ddagger$ & Arxiv'24 & 7B & 64 & 64.4 & 74.6 & 62.9 & 55.7 \\
        \rowcolor{cyan!15} \textbf{LLaVA-Video w/Ours} & \textbf{-} & \textbf{7B} & \textbf{Avg 56} & \textbf{66.2} & \textbf{77.6} & \textbf{64.4} & \textbf{56.6} \\
        \bottomrule
        \end{tabular}
    }
    \vspace{0.2cm}
    \caption{\textbf{Quantitative comparison on Video-MME.} Baseline methods reproduced with official code are marked with $\ddagger$. Best results in \textbf{bold}, second best \underline{underlined}.}
    \label{tab:sota_video_mme}
    \vspace{-0.8cm}
\end{table*}

\subsection{Benchmark Results}
\textcolor{black}{Video-EM uses keyframes as input but departs from prior \emph{frame-centric} pipelines by organizing and compressing them into \emph{event-level} episodic memories. Accordingly, we focus on comparisons with strong keyframe selection baselines to demonstrate the advantage brought by event-centric structuring and refinement under the same keyframe evidence budget.}
We mainly report results using Qwen2-VL~\cite{wang2024qwen2}, Qwen2.5-VL~\cite{bai2025qwen2}, LLaVA-OV~\cite{li2025llavaonevision}, and LLaVA-Video~\cite{zhang2024video} as backbones to show the effectiveness and versatility of Video-EM. 

\noindent \textbf{Comparison with State-of-the-Art Methods.} 
We evaluate the effectiveness of Video-EM across four representative long-form video benchmarks: Video-MME, LVBench, HourVideo, and Egoschema.
As shown in Table~\ref{tab:sota_video_mme}, compared with efficient keyframe sampling methods on Video-MME, such as AKS~\cite{tang2025adaptive}, BOLT~\cite{liu2025bolt},
 and Q-Frame~\cite{zhang2025q}, and FOCUS~\cite{zhu2025focus}, Video-EM consistently outperforms existing training-free approaches and achieves highly competitive results among open-source models on all four benchmarks.
Notably, as illustrated in Table~\ref{tab:sota_other_benchmarks}, our method improves performance by 7\% on  LVBench and 3\% on HourVideo, while requiring substantially fewer frames (27 \textit{vs.} 64 on LVBench and 30 \textit{vs.} 64 on HourVideo).
Furthermore, on the Egoschema benchmark (Table~\ref{tab:combined_analysis} (a)), our framework shows notable effectiveness compared to the baseline, improving performance to 65.6\% and 64.4\%, respectively, while reducing the number of frames used from 16 to 9.
These results highlight the strong generalization capabilities of Video-EM, significantly outperforming existing open-source state-of-the-art methods and underscoring its adaptability to complex, egocentric video scenarios.
Overall, these results demonstrate that Video-EM effectively mitigates the limitations of previous keyframe-based methods, and is both widely applicable and robust across a range of long-video understanding tasks.

\begin{figure*}[t]
    \centering
    \includegraphics[width=1\linewidth]{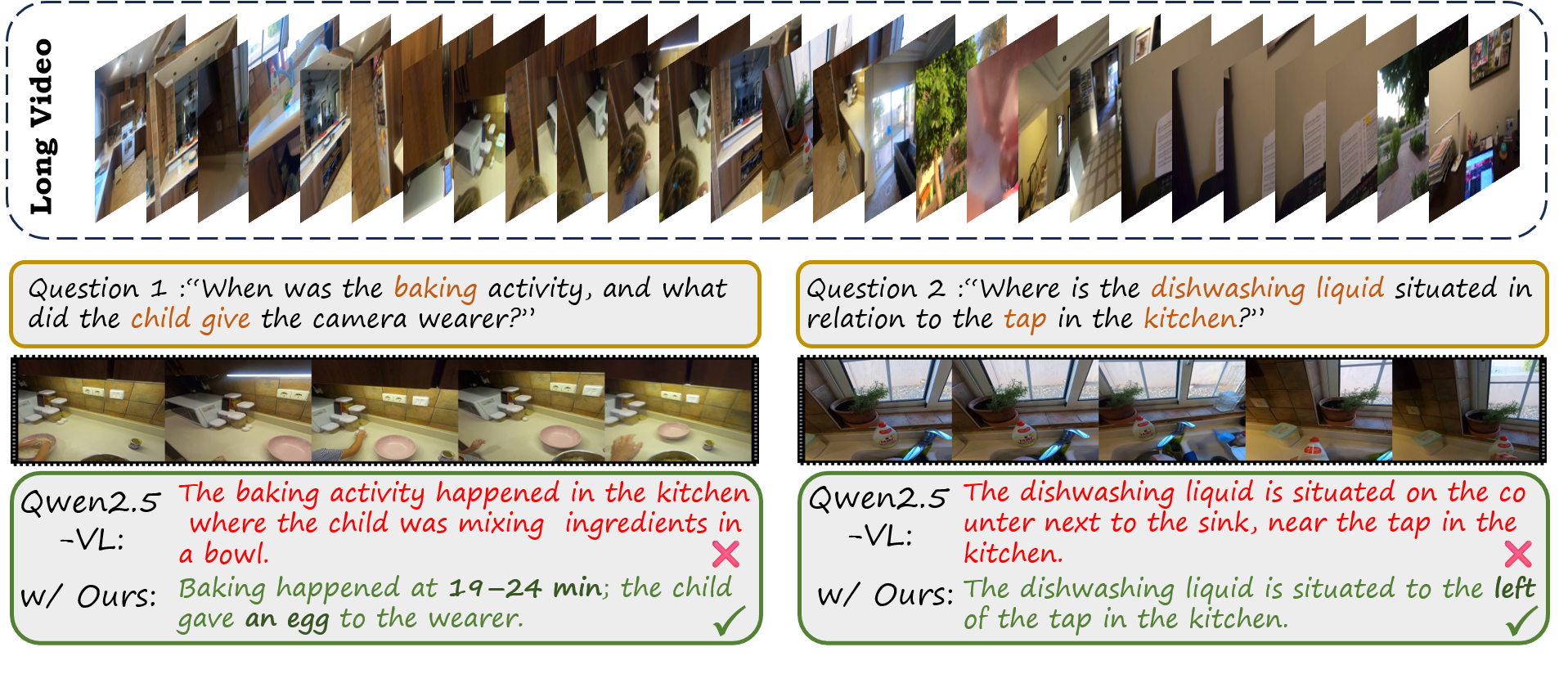}
    \vspace{-0.8cm}
    \caption{Qualitative examples from HourVideo comparing our model (\textcolor{MyForestGreen}{green}) with Qwen2.5-VL (\textcolor{red}{red}). Frames are manually selected to highlight query-relevant events.}
    \vspace{-0.3cm}
\label{fig:qualitative}
\end{figure*}

\begin{table*}[t]
    \centering
    \small
    
    \begin{minipage}[t]{0.49\linewidth}
        \centering
        \caption*{\small{\textbf{LVBench}}} 
        \vspace{-0.3cm}
        \resizebox{0.95\linewidth}{!}{
            \renewcommand{\arraystretch}{1.3}
            \setlength{\tabcolsep}{3pt}
            \begin{tabular}{l c c c c}
            \toprule
            \multirow{2}{*}{Model} & \multirow{2}{*}{Venue} & \multirow{2}{*}{LLM Size} & \multirow{2}{*}{Frames} & Overall \\
            \cmidrule(lr){5-5} 
             & & & &68 min \\
            \midrule
            GPT-4o & - & - & - & 34.7 \\
            Gemini 1.5 pro & - & - & - & 33.1 \\
            \midrule
            LLaVA-Video & ArXiv'24 & 7B & 180 & 41.5 \\
            Video-XL & CVPR'25 & 7B & - & 36.8 \\
            Video-XL-2 & Arxiv'25 & 8B & 1fps & 48.4 \\
            VAMBA & CVPR'25 & 10B & 1024 & 42.1 \\
            \textbf{Qwen2-VL} & ArXiv'24 & 72B & 1fps & 44.0 \\ 
            \midrule
            Qwen2-VL$^\ddagger$ & ArXiv'24 & 7B & 32 & 38.4 \\
            Qwen2-VL$^\ddagger$ & ArXiv'24 & 7B & 64 & 39.7 \\
            \rowcolor{cyan!15} \textbf{Qwen2-VL w/Ours} & Arxiv'25 & 7B & Avg 27 & \underline{45.2} \\  
            \midrule
            Qwen2.5-VL$^\ddagger$ & ArXiv'25 & 7B & 32 & 36.6 \\  
            Qwen2.5-VL$^\ddagger$ & ArXiv'25 & 7B & 64 & 38.7 \\
            \rowcolor{cyan!15} \textbf{Qwen2.5-VL w/Ours} & - & 7B & Avg 27 & \textbf{45.7} \\  
            \bottomrule
            \end{tabular}
        }
    \end{minipage}
    \hfill 
    \begin{minipage}[t]{0.49\linewidth}
        \centering
        \caption*{\small{\textbf{HourVideo}}} 
        \vspace{-0.3cm}
        \resizebox{0.95\linewidth}{!}{
            \renewcommand{\arraystretch}{1.38}
            \setlength{\tabcolsep}{3pt}
            \begin{tabular}{l c c c c}
            \toprule
            \multirow{2}{*}{Model} & \multirow{2}{*}{Venue} & \multirow{2}{*}{LLM Size} & \multirow{2}{*}{Frames} & Overall \\
            \cmidrule(lr){5-5} 
             & & & & 47 min \\
            \midrule
            GPT-4 & - & - & 0.5fps & 25.7 \\
            Gemini 1.5 Pro & - & - & 0.5fps & 37.3 \\
            \midrule
            InternVL3 & ArXiv'25 & 7B & 32 & 30.1 \\
            LongVU & ICML'25 & 7B & 1fps & 30.8 \\
            VideoLLaMA3 & ArXiv'25 & 7B & 32 & 31.0 \\
            VAMBA & CVPR'25 & 10B & 128 & 33.6 \\
            \midrule
            Qwen2-VL$^\ddagger$ & ArXiv'24 & 7B & 32 & 30.7 \\
            Qwen2-VL$^\ddagger$ & ArXiv'24 & 7B & 64 & 32.4 \\   
            \rowcolor{cyan!15} \textbf{Qwen2-VL w/Ours} & - & 7B & Avg 30 & \underline{34.8} \\   
            \midrule
            Qwen2.5-VL$^\ddagger$ & ArXiv'25 & 7B & 32 & 31.1 \\ 
            Qwen2.5-VL$^\ddagger$ & ArXiv'25 & 7B & 64 & 32.8 \\
            \rowcolor{cyan!15} \textbf{Qwen2.5-VL w/Ours} & - & 7B & Avg 30 & \textbf{35.1} \\ 
            \bottomrule
            \end{tabular}
        }
    \end{minipage}
    \vspace{0.2cm}
    \caption{\textbf{Quantitative comparison on LVBench (left) and HourVideo (right).} Baseline methods reproduced with official code are marked with $\ddagger$.}
    \label{tab:sota_other_benchmarks}
    \vspace{-1.0cm}
\end{table*}

\noindent \textbf{Improved Results on Off-the-Shelf VLMs.} 
As shown in Table~\ref{tab:sota_video_mme},  our Video-EM consistently improve the performance of off-the-shelf Vision Language Models (VLMs) without requiring training. When applied to mainstream models like Qwen2-VL~\cite{wang2024qwen2} and Qwen2.5-VL~\cite{bai2025qwen2} across four long video benchmarks, we observe significant gains. 
\textcolor{black}{Compared with other keyframe selection methods, Video-EM significantly outperforms them under the same backbone (Qwen2-VL), achieving gains of +1.3\% over AKS~\cite{tang2025adaptive}, +2.9\% over Q-Frame~\cite{zhang2025q}, and +1.7\% over BOLT~\cite{liu2025bolt}, while using fewer frames.}
This fully demonstrates its effectiveness, with qualitative examples provided in Figure~\ref{fig:qualitative}.}

\begin{table*}[t]
    \centering
    \small
    \setlength{\tabcolsep}{3pt} 
    \renewcommand{\arraystretch}{1.1} 

    \begin{minipage}[t]{0.54\linewidth}
        \vspace{0pt} 
        \caption*{\textbf{(a) Performance Comparison}}
        \vspace{-0.3cm}
        
        \resizebox{\linewidth}{!}{
            \begin{tabular}{lcccc}
            \toprule
            \multirow{2}{*}{Model} & \multirow{2}{*}{Venue} &  \multirow{2}{*}{LLM Size} & \multirow{2}{*}{Frames}  & Overall  \\
            \cmidrule(lr){5-5} 
             & & & & 3 min \\
            \midrule
            PLLaVA & ArXiv'24 & 7B & 16 & 45.6 \\
            VideoLLaMA2 & ArXiv'24 & 7B & 8 & 53.1 \\
            VideoChat2 & CVPR'24 & 7B & 16 & 54.4\\
            VideoLLaVA & EMNLP'24 & 7B & 8 & 40.2\\
            \midrule
            LLoVi & EMNLP'24 & GPT-3.5 & 90 & 57.6 \\
            VideoAgent & ECCV'24 & GPT-4 & 8.4 & 60.2\\
            VideoLLaMa & ICCV'25 & 7B & 8 & 53.8\\
            VideoTree & CVPR'25 & GPT-4 & 63.2 & \textbf{66.2} \\
            BOLT & CVPR'25 & 7B & 16 & 61.8\\
            MVU & ICLR'25 & 7B & 16 & 60.2\\
            \midrule
            Qwen2-VL$^\ddagger$  & ArXiv'24 & 7B  & 8 & 59.6\\
            Qwen2-VL$^\ddagger$ & ArXiv'24 & 7B  & 16 & 61.2 \\
            \rowcolor{cyan!15} \textbf{Qwen2-VL w/ Ours}  & - & 7B & Avg 9 & \underline{65.6} \\   
            \cdashline{1-5} 
            
            Qwen2.5-VL$^\ddagger$ & ArXiv'25 & 7B  & 8 & 56.4 \\
            Qwen2.5-VL$^\ddagger$ & ArXiv'25 & 7B  & 16 & 60.2 \\
            \rowcolor{cyan!15} \textbf{Qwen2.5-VL w/ Ours} & - & 7B  & Avg 9 & 64.4 \\ 
            \midrule
            LLaVA-OV$^\ddagger$ & ArXiv'24 & 7B & 16 & 61.4 \\
            \rowcolor{cyan!15} \textbf{LLaVA-OV w/ Ours} & - & 7B & Avg 9 & 65.2\\
            \bottomrule
            \end{tabular}
        }
    \end{minipage}
    \hfill 
    \begin{minipage}[t]{0.44\linewidth}
        \vspace{0pt} 
        
        \caption*{\textbf{(b) Ablations of components}} 
        \vspace{-0.3cm}
        \resizebox{\linewidth}{!}{
            \setlength{\tabcolsep}{8pt} 
            \begin{tabular}{cccc}
            \toprule
            ID & Settings & Frames & Acc (\%) \\
            \midrule
            1 & \textbf{Full} & \textbf{Avg 9} & \textbf{64.4} \\
            \midrule
            2 & w/o EMC  & Avg 8 & 59.0 \\
            3 & w/o EES  & Avg 8 & 63.2  \\
            4 & w/o DSN & Avg 9 & 59.0 \\
            5 & w/o DSR & Avg 9 & 62.8 \\
            \midrule
            6 & w/o CoT & Avg 41 & 62.8 \\
            \bottomrule
            \end{tabular}
        }
        
        \vspace{0.3cm} 
        
        \caption*{\textbf{(c) Computational cost analysis}}
        \vspace{-0.2cm}
        \resizebox{\linewidth}{!}{
            \setlength{\tabcolsep}{2pt}
            \renewcommand{\arraystretch}{1.4}
            \begin{tabular}{lcccc}
            \toprule
            Method & Acc (\%) &Preprocess & Time (s)  & FLOPs (T) \\
            \midrule
            Baseline & 59.6 &-- & -- & -- \\
            + MSR     & 60.5 &CLIP & 0.35 & 20.7 \\
            + EES     & 61.1 &TransNet v2 & 0.11 & $4.1\times10^{-3}$ \\ 
            + EMC      & 64.2&Qwen & 1.56 & 140 \\
            + CoT     & 65.6 &Qwen & 2.63 & 132.1 \\
            \bottomrule
            \end{tabular}
        }
    \end{minipage}
    
    \vspace{0.2cm} 
    \caption{\textbf{Comprehensive analysis on EgoSchema.} \textbf{(a)} Comparison with state-of-the-art methods. \textbf{(b)} Ablation study of different Video-EM components. \textbf{(c)} Computational ablation of Video-EM. ``Preprocess'' indicates the auxiliary models used at each stage. Time (s) denotes the average inference time per video on EgoSchema. FLOPs are measured based on the actual number of frames processed by each component.}
    \label{tab:combined_analysis}
    \vspace{-1.0cm} 
\end{table*}

\begin{figure}[t]
  \vspace{-2pt} 
      \centering
    \includegraphics[width=0.85\linewidth]{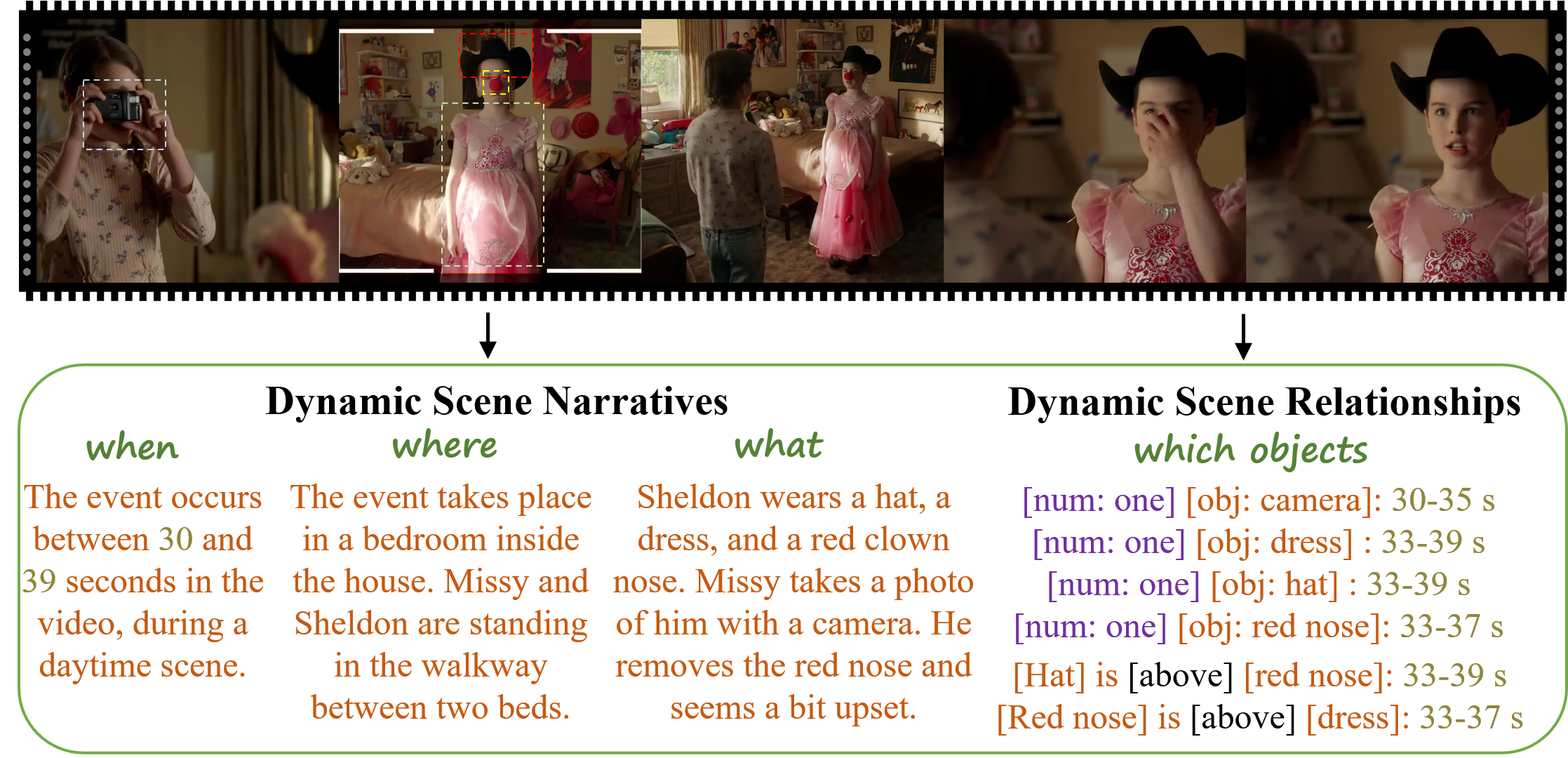}
    \vspace{-0.2cm}
    \caption{Qualitative examples of our Episodic Memory, composed of Dynamic Scene Narratives  and Dynamic Scene Relationships.}
    \label{fig:em}
    \vspace{-0.4cm}
\end{figure}

\begin{figure}
    \centering
    \includegraphics[width=0.8\linewidth]{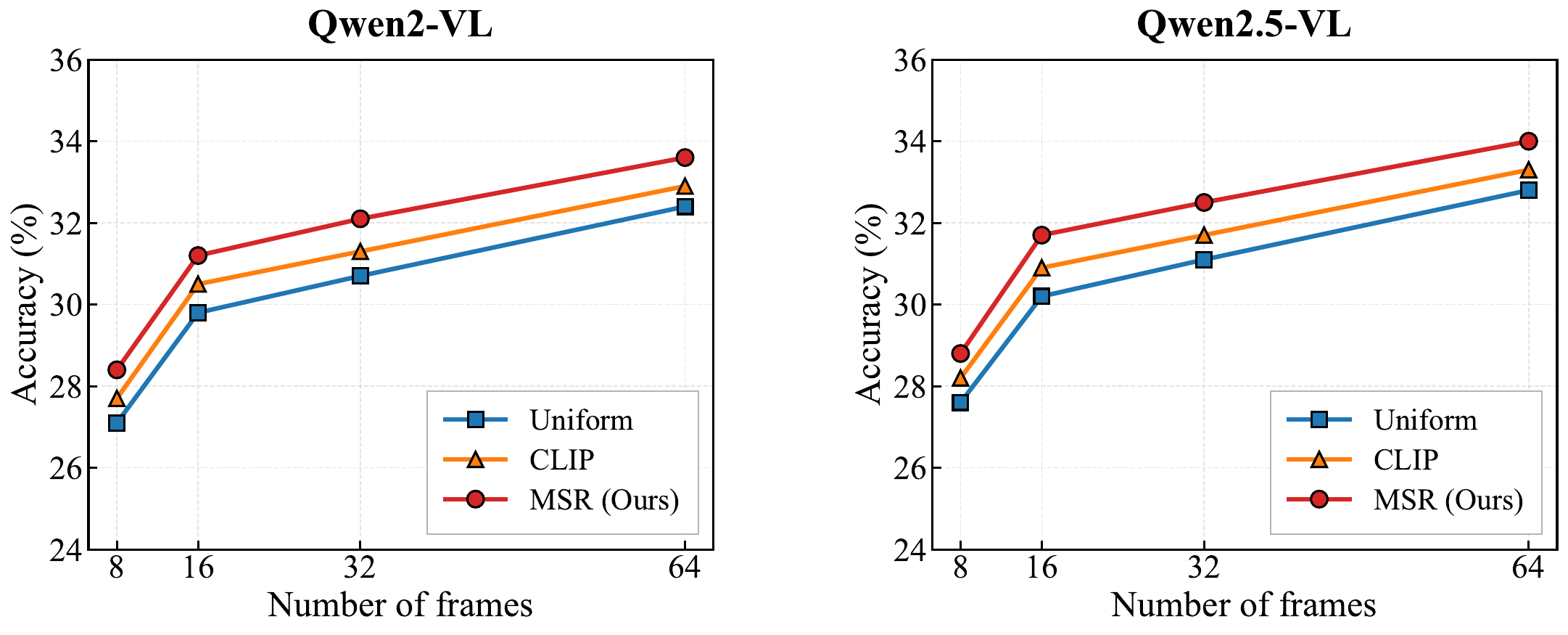}
    \vspace{-0.3cm}
    \caption{Ablation study of the number of frames under different sampling strategies on the HourVideo dataset.}
    \label{fig:frames}
    \vspace{-0.6cm}
\end{figure}

\vspace{-0.4cm}
\subsection{Ablation Studies}
\vspace{-0.1cm}
We ablate on Egoschema using Qwen2.5-VL as the MLLM backbone to assess the contributions of Video-EM model variants.


\noindent \textbf{Ablation on Core Components.}
We evaluate the individual contributions of Video-EM’s core components: Episodic Memory Construction (EMC), Event Expansion \& Segmentation (EES), Dynamic Scene Narratives (DSN), Dynamic Scene Relationships (DSR), and the Chain-of-Thought (CoT) reasoning module. 
Specifically, we process 128 frames as input and select 16 frames as initial keyframes for the subsequent EES execution. As shown in Table~\ref{tab:combined_analysis} (b), the exclusion of EMC leads to a performance drop from 64.4\% to 59.0\% (ID = 1 \textit{vs.} ID = 2). This result highlights the crucial role of structured episodic memory in supporting high-level video reasoning. The architecture of our EMC is illustrated in Figure~\ref{fig:em}.
Within EMC, eliminating the EES module (ID = 3) results in reliance solely on similarity-based keyframe retrieval. This neglects surrounding contextual information and often captures only partial events, leading to suboptimal coverage and a large performance drop.
Furthermore, removing the CoT module (ID = 6) dramatically increases the number of selected frames, from 9 to 41, while the accuracy decreases from 64.4\% to 62.8\%. This suggests that overloading the model with redundant visual inputs can degrade reasoning efficiency and accuracy. The result highlights that selecting a minimal yet semantically rich subset of frames is effective than processing redundant visual content.

\noindent \textbf{Effect of Keyframe Selection.}
We evaluate three keyframe selection strategies: Uniform sampling, CLIP-based similarity retrieval, and Multi-grained Semantic Retrieval (Figure~\ref{fig:frames}), with all subsequent modules disabled. Compared to the former two, MSR decomposes each query into multi-granular semantic cues to retrieve contextually relevant frames more precisely, resulting in a substantial accuracy gain at the cost of using more keyframes.

\noindent \textbf{Effect of Varying Keyframes.}
We study the effect of temporal granularity by varying the frame budget \emph{under the full model setting} on two long-form video benchmarks, Video-MME and LVBench. As shown in Table~\ref{tab:hyperparams} (left), performance improves monotonically within the evaluated range as more frames are provided, highlighting the importance of denser temporal sampling. Additional frames offer finer motion cues and better coverage of event boundaries, enabling more reliable reasoning over long-form narratives.

\begin{table}[t]
    \centering 
    \small
    \begin{minipage}[c]{0.45\linewidth} 
        \raggedleft 
        \renewcommand{\arraystretch}{1.15} 
        \setlength{\tabcolsep}{4pt} 
        \begin{tabular}{l c c}
        \toprule
        \textbf{Dataset} & \textbf{Frames} & \textbf{Acc (\%)} \\
        \midrule
        \multirow{3}{*}{V-MME} 
            & Avg 7   & 56.7 \\
            & Avg 14  & 59.4 \\
            & Avg 28  & \textbf{62.0} \\
        \midrule
        \multirow{3}{*}{LVBench} 
            & Avg 7   & 37.3 \\
            & Avg 13  & 40.6 \\
            & Avg 27  & \textbf{45.7} \\
        \bottomrule
        \end{tabular}
    \end{minipage}
    \hspace{0.3cm} 
    %
    \begin{minipage}[c]{0.45\linewidth} 
        \raggedright 
        \includegraphics[height=3.8cm, keepaspectratio]{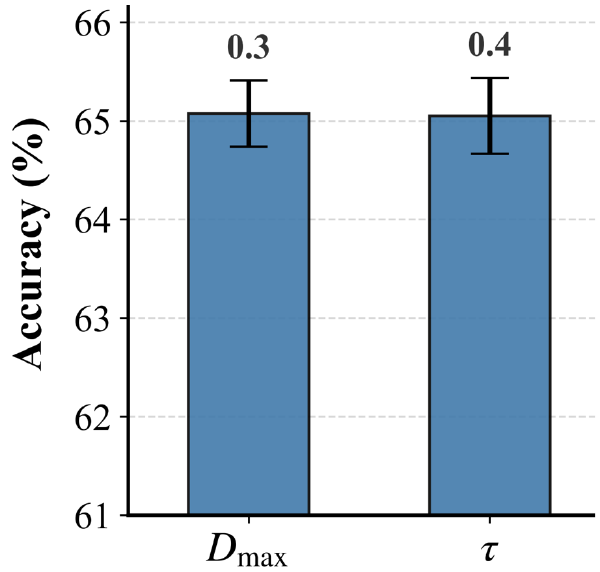}
    \end{minipage}
    
    \vspace{0.1cm}
    \caption{\emph{Left}: Key-event selection performance under varying number of keyframes.
\emph{Right}: CoT hyperparameter analysis with $D_{\text{max}}\!\in\![1,3]$ and $\tau\!\in\![0.5,1.0]$.}
    \label{tab:hyperparams}
    \vspace{-0.8cm}
\end{table}


\begin{table}[t]
    \centering
    \small
    \begin{minipage}[t]{0.45\linewidth}
        \centering
        \caption*{\textbf{(a) Modality Contribution}} 
        \vspace{-0.3cm}
        \setlength{\tabcolsep}{5pt}
        \renewcommand{\arraystretch}{1.4} 
        \begin{tabular}{ccc} 
            \toprule
            \textbf{Frames} & \textbf{EM Text} & \textbf{Acc (\%)} \\
            \midrule
            \checkmark &  & 61.1 \\ 
             & \checkmark & 63.8 \\ 
            \rowcolor{gray!15} \checkmark & \checkmark & \textbf{65.6} \\
            \bottomrule
        \end{tabular}
    \end{minipage}
    \hfill 
    \begin{minipage}[t]{0.52\linewidth}
        \centering
        \caption*{\textbf{(b) Query Components}} 
        \vspace{-0.3cm}
        \setlength{\tabcolsep}{6pt}
        \renewcommand{\arraystretch}{1.16}
        \begin{tabular}{cccc}
            \toprule
            $q_o$ & $q_s$ & $q_c$ & \textbf{Acc (\%)} \\
            \midrule
            \checkmark &  &  & 63.9 \\
            \checkmark & \checkmark &  & 64.4 \\
            \checkmark &  & \checkmark & 64.8 \\
            \rowcolor{gray!15} \checkmark & \checkmark & \checkmark & \textbf{65.6} \\
            \bottomrule
        \end{tabular}
    \end{minipage}
    \vspace{0.1cm}
    \caption{\textbf{Ablation Studies.} We investigate the contribution of different input modalities (Left) and the impact of query decomposition components (Right): original query ($q_o$), object-level query ($q_s$), and scene-level query ($q_c$).}
    \label{tab:ablation_combined_1}
    \vspace{-1cm} 
\end{table}


\noindent \textbf{Cost Consumption Analysis.}
To evaluate the computational cost of our method, we initially sample 128 frames and select 9 frames for the final Video-LLM inference. MSR denotes Multi-grained Semantic Retrieval, and EES denotes Event Expansion and Segmentation.
As shown in Table~\ref{tab:combined_analysis} (c), enabling additional modules consistently improves performance, with the full pipeline achieving 65.6\% accuracy (+6.0\% over the baseline). Although adding more modules increases computation, the end-to-end runtime remains moderate (4.75 $s$), indicating a favorable trade-off between cost and accuracy. Notably, lightweight components such as MSR and EES deliver early gains with minimal overhead.


\noindent \textbf{Ablation on the CoT Module.}
We investigate the influence of two key hyperparameters in the CoT module: the maximum reasoning depth ($D_{\text{max}}$) and the confidence threshold ($\tau$), which govern the iterative selection and refinement of episodic memories. As summarized in Table~\ref{tab:hyperparams} (right), Video-EM maintains consistently strong accuracy against hyperparameter variations, demonstrating that its performance is highly robust and largely insensitive to hyperparameter tuning. 
This resilience underscores the practicality of our framework, as it can be readily applied without extensive parameter optimization.

\noindent \textbf{Ablation on the input type.}
To locate the source of Video-EM’s gains, we further disentangle the inputs to the Video-LLM: {frames only}, {EM text only}, and {both}.
As shown in Table~\ref{tab:ablation_combined_1} (left), frames alone achieve 61.1\%, suggesting that sparse keyframes may provide incomplete or weakly grounded evidence for long-horizon queries, where cues are distributed across time and entangled with repetitive scenes.
Using only EM text improves to 63.8\%, indicating that structured \emph{when/where/what}-style abstraction acts as a strong prompt-level constraint: it compresses long-range context into \emph{temporally coherent} salient events, highlights entities and relations, and helps the LLM follow a more disciplined reasoning trajectory without raw visuals.
Combining frames with EM text yields the best result (65.6\%), confirming complementarity: frames retain fine-grained visual details, while EM text provides compact episodic structure and {explicit} grounding cues, enabling more stable complex spatio-temporal reasoning.

\noindent \textbf{Ablation on query decomposition in the MSR module.}
We ablate the query decomposition used in the MSR module by comparing the original query $q_o$ with additional object-level $q_s$ and scene-level $q_c$ components (Table~\ref{tab:ablation_combined_1}, right). 
Using only $q_o$ reaches 63.9\%. Adding $q_s$ improves to 64.4\%, while adding $q_c$ yields 64.8\%, indicating that object- and scene-centric cues benefit retrieval from complementary perspectives. 
Using all components ($q_o{+}q_s{+}q_c$) achieves the best performance (65.6\%), validating that multi-grained decomposition improves evidence coverage for long-form VideoQA.

\begin{figure*}[t]
    \centering
    \vspace{-0.3cm}
    \includegraphics[width=1\linewidth]{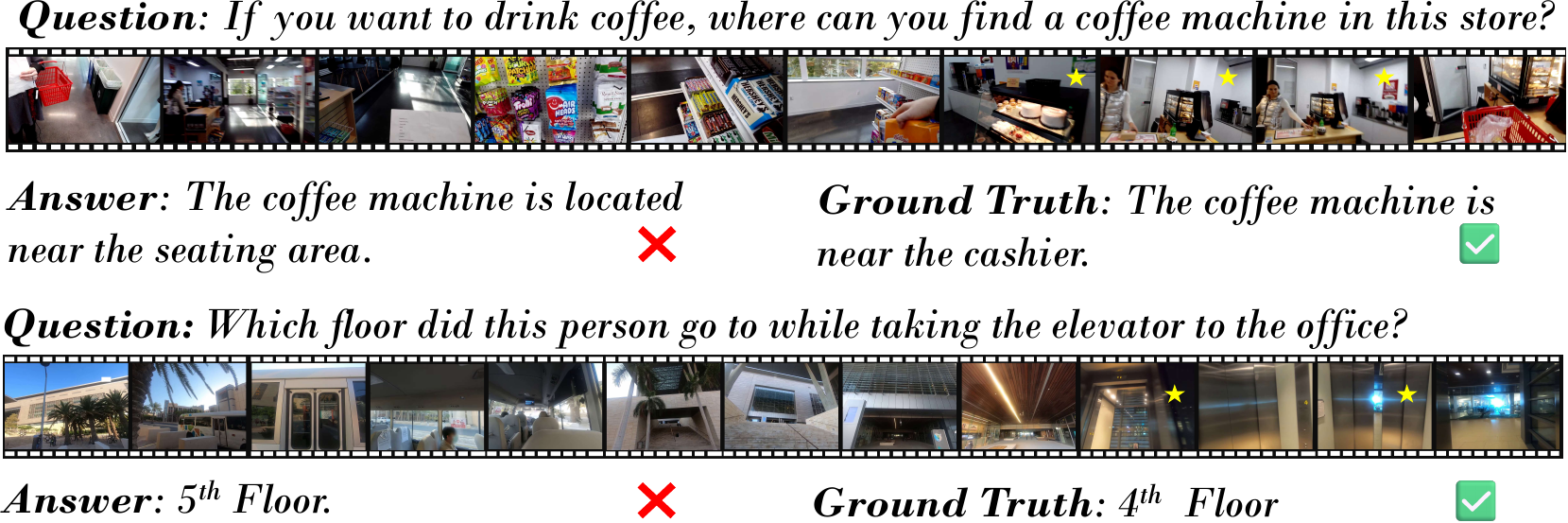}
    \vspace{-0.5cm}
    \caption{Two representative failure modes of Video-EM when using Qwen2.5-VL-7B as the baseline for key-event selection are shown.  Yellow stars mark manually annotated frames that are most informative for the query.
    }
    \label{fig:failure_case}
    \vspace{-0.6cm}
\end{figure*}

\vspace{-0.3cm}
\section{Failure Case}
\vspace{-0.2cm}
To better understand Video-EM, we analyze two representative failure patterns on Qwen2.5-VL-7B when using Video-EM for key event selection (Figure~\ref{fig:failure_case}). 
\textbf{(1) Limited instance-level grounding in MLLMs.} For ``If you want to drink coffee, where can you find a coffee machine in this store?'', a visually similar coffee machine recurs across near-duplicate moments; despite coherent episodic memory, the MLLM may conflate instances and ground the answer to the wrong occurrence, reflecting limited fine-grained event understanding in MLLMs. 
\textbf{(2) Needle-in-a-haystack temporal evidence.} In a 50-minute video with frequent scene transitions, the query hinges on a 1-2 second decisive cue. Video-EM may retrieve temporally nearby evidence, but reliable inference remains difficult for current MLLMs when the key signal is fleeting and highly localized.



\vspace{-0.3cm}
\section{Conclusion}
In this paper, we present Video-EM, a training-free framework for long-form video understanding in Video-LLMs. Instead of treating keyframes as isolated snapshots, Video-EM organizes them into temporally ordered events, encodes these events as episodic memories, and further employs a reasoning-driven refinement loop to produce a minimal yet sufficient \emph{event timeline} for downstream answering, suppressing redundancy and noise without retraining or architectural changes. 
This design enables stronger spatio-temporal reasoning while remaining efficient. Video-EM is plug-and-play, integrating with diverse Video-LLM backbones without any architectural changes. Extensive experiments across mainstream long-form video benchmarks demonstrate substantial performance gains.

\vspace{-0.4cm}
\bibliographystyle{splncs04}
\bibliography{main}
\end{document}